\documentclass[]{article}

\usepackage[utf8]{inputenc}
\usepackage{amssymb}
\usepackage{hyperref}
\usepackage{amsmath}
\usepackage{makecell}
\usepackage{graphicx, color}
\usepackage[sort&compress,numbers,square]{natbib}
\usepackage{multicol}
\usepackage{subcaption}

\title{Automatic Translation of Hate Speech to Non-hate Speech in Social Media Texts}

\author{Yevhen Kostiuk$^1$ \and Atnafu Lambebo Tonja$^1$ \and Grigori Sidorov$^1$ \and Olga Kolesnikova$^1$}
\date{
	$^1$Instituto Polit\'ecnico Nacional, Centro de Investigaci\'{o}n en Computaci\'{o}n, Av. Juan de Dios Batiz, s/n, 07320, Mexico City, Mexico \\ \texttt{kosteugeneo@gmail.com, atnafu.lambebo@wsu.edu.et, sidorov@cic.pin.mx, kolesolga@gmail.com}\\%
}

\begin{document}

\flushbottom
\maketitle
\thispagestyle{empty}

\section*{Disclaimer}
This paper includes examples of hate speech, which may include offensive language and discriminatory content. These examples are used for research purposes only and do not reflect the views or beliefs of the authors. We acknowledge the potential harm that such language may cause and encourage the responsible use of language in all contexts.

\begin{abstract}
In this paper, we investigate the issue of hate speech by presenting a novel task of translating hate speech into non-hate speech text while preserving its meaning. As a case study, we use Spanish texts. We  provide a dataset and several baselines as a starting point for further research in the task. We evaluated our baseline results using multiple metrics, including BLEU scores. The aim of this study is to contribute to the development of more effective methods for reducing the spread of hate speech in online communities.
\end{abstract}

\section*{Introduction}

Nowadays, social media sites play an important role in society, and their impact cannot be overestimated. People with different backgrounds around the globe can share their opinions and communicate via social media networks anonymously. However, a disadvantage of social media is the ability to send inappropriate messages to anyone without consequence, including hate messages. Hate speech is defined as a speech or writing that aims to disparage a person or a group of people on the grounds of race, religious beliefs, sexual orientation, nationality, or similar characteristics~\cite{thota2018fake}. Hate speech can be disseminated through social media platforms such as Twitter and Facebook, according to~\cite{allcott2017social}. This paves the way for an infinite number of potential outcomes, while at the same time mitigating a considerable risk, given that users are vulnerable to a wide variety of dangers and assaults, one of which is hateful speech. Hate may inflict victims both immediate and long-term harm. It affects public opinion formation, creates harmful stereotypes, and targets individuals~\cite{ortiz2019detection}. It may even be the cause of suicides and self-harm in some cases~\cite{10.1001/jamanetworkopen.2021.25860, saha2019prevalence}. Thus, it is very important to stop its spread. Hate messages can be deliberate and not deliberate, the latter is the case when people do not intend to harm anyone in any way but they sound rude. Such users can be carried away in an online argument or have a bad time. In such situations, an algorithm that can "translate" hate speech into a more acceptable text can be very helpful. 
In this paper, we propose a new hate speech translation task and use a new Spanish dataset as a case study. We ran experiments on our dataset using three pre-trained language models to investigate the dataset's usability.

One approach to the task we defined here is the use of natural language processing (NLP) tools to automatically translate hate speech into non-hateful language. The idea behind hate speech to non-hate speech translation is to intervene in online conversations and reduce the impact of hate speech by providing an alternative and non-offensive version of the text. This approach involves automatically identifying instances of hate speech and replacing them with non-hateful language while preserving the overall meaning and structure of the text.

The contributions of our research are as follows:
\begin{itemize}
    \item We introduce a new task for hate speech to non-hate speech translation.
    \item We introduce the first hate speech translation dataset for the Spanish language, as our case study.
    \item We present the results of our experiments on hate speech translation.
    \item We open-source our dataset and code for researchers interested in this area.
\end{itemize}
The code and dataset is available here\footnote{CURRENTLY HIDDEN FOR REVIEW}
The rest of the article is organized as follows. In Section~\hyperlink{sec:background}{\textbf{Background of the study}}, previous research related to our study is described. Section~\hyperlink{sec:task}{\textbf{Task}} presents the proposed task. Section~\hyperlink{sec:dataset}{\textbf{Dataset}} gives statistics of the dataset. Section~\hyperlink{sec:method}{\textbf{Baseline Methodology}} presents baseline models used for the task. Section~\hyperlink{sec:eval}{\textbf{Testing and Evaluation}} explains the dataset split and the metrics used to evaluate our system. Section~\hyperlink{sec:exper}{\textbf{Experiments and Results}} presents the experimental results. Finally, Section~\hyperlink{sec:conc}{\textbf{Conclusion}} concludes the paper and sheds some light on possible future work.

\section*{Background of the study} \label{background}\hypertarget{sec:background}{}

Hate speech translation (more details are provided in Section~\ref{task}) is related to the following machine learning tasks: hate speech detection, paraphrasing, and machine translation. The goal of these tasks is to generate semantically equivalent text in another language or with different wording while preserving the original meaning without any changes to it. In contrast, we propose a similar yet different approach. In our task of hate speech to non-hate speech translation, the algorithm will decide, which meanings in the input text are harmful and will be removed, and which ones will be preserved in the output in the same language. The semantic analysis required to fulfill it must be deeper than the case of paraphrasing. In other words, not only the same message in the original text should be conveyed but also it should be rewritten in a way that is socially acceptable and free of harmful or offensive language. It is also different from machine translation tasks because, in machine translation, we usually translate one language to another~\cite{belay2022effect}. As such, the task presents unique challenges that have not been fully explored in previous research. To the best of our knowledge, the proposed task in this work has not yet been introduced for research in the area of hate speech processing to automatically identify and replace hate speech with non-hate speech, and no dataset for
this specific task is available up to date.
So, the task we propose of hate speech to non-hate speech translation can be seen as a form of paraphrasing, on the one hand, and the other hand, as translation, where the goal is to rephrase or “translate” a text to remove hate speech while preserving its original meaning. To see some differences and similarities between our task of paraphrasing and translation, we now review the latter two NLP areas.

Paraphrasing is a restatement of a passage using different and synonym words to express the same message for various purposes, including text simplification, data augmentation, and plagiarism detection, among others~\cite{socher2011dynamic, okur2022data}.

Paraphrasing can be approached using rule-based methods, such as synonym replacement and syntactic transformations, or using machine learning-based methods, such as neural sequence-to-sequence models~\cite{madnani2010generating, iyyer2018adversarial}. In recent years, neural models have become increasingly popular for paraphrasing, as they can learn to generate diverse and fluent paraphrases by training on large corpora of sentence pairs.

A common approach for generating paraphrases using neural models is to use an encoder-decoder architecture with attention mechanisms~\cite{egonmwan-chali-2019-transformer-seq2seq}. Given a source sentence, the encoder produces a fixed-length vector representation, which is used by the decoder to generate a paraphrase. Several variations of this architecture have been proposed, such as using pre-trained language models and incorporating lexical and semantic constraints~\cite{witteveen-andrews-2019-paraphrasing, wieting-etal-2019-simple}.

Paraphrasing has also been studied in the context of specific domains and tasks, such as text simplification~\cite{maddela2020controllable, yimam2018par4sim, xu2016optimizing}, question answering~\cite{fader2013paraphrase, kacupaj2021paraqa, abujabal2018comqa}, style transfer~\cite{shen2017style}, information extraction~\cite{shinyama2003paraphrase}, and semantic parsing~\cite{berant2014semantic, su2017cross} tasks. Our proposed task of hate speech to non-hate speech translation can be seen as a form of paraphrasing, where the goal is to rephrase a text to remove hate speech while preserving its original meaning. Despite the progress in paraphrasing models, there are still challenges in generating high-quality paraphrases, such as avoiding semantic drift, handling rare and out-of-vocabulary words, and ensuring grammatical correctness~\cite{zhou-bhat-2021-paraphrase}. Addressing these challenges remains an active area of research.

Machine translation (MT) is a branch of NLP that studies how to utilize computer software to automatically translate spoken or written text from one language to another without the need for human intervention~\cite{belay2022effect}. In recent years, the development of deep neural networks and sequence-to-sequence models has led to significant improvements in the quality of MT systems~\cite{wu2016googles, vaswani2017attention}. These models can learn to map sequences of words from a source language(s) to a target language(s) by processing large amounts of bilingual training data. One of the most popular approaches to MT is neural machine translation (NMT), which uses a neural network to model the probability distribution over the target language given the source sentence~\cite{sutskever2014sequence}. NMT has been applied to a wide range of languages and domains, achieving state-of-the-art results on many benchmark datasets~\cite{bojar2016findings}. In addition to traditional translation tasks, recent research has explored the use of MT for other language-related tasks, such as text simplification~\cite{xu2016optimizing} or style transfer~\cite{prabhumoye2018style}. Our proposed task of hate speech translation can be seen as another example of leveraging MT to address societal challenges. 
Many researchers explored machine translation tasks for different languages~\cite{sen2018iitp,choudhary2018neural,saini2018neural,belay2022effect,tonja2023low}, using different approaches~\cite{lopez2008statistical,stahlberg2020neural,artetxe2017unsupervised, currey2019zero, tonja2022improving, tonja2023low} and in different directions~\cite{johnson2017google,aharoni2019massively,tan2019multilingual}. Despite the recent progress, MT systems still face significant challenges, such as handling rare words and out-of-vocabulary tokens, modeling long-term dependencies, and capturing complex linguistic phenomena~\cite{koehn2017six, bahdanau2016neural}.

Hate speech is a complex NLP concept. Sometimes it is not easy to decide whether a given text can be classified as hate or not. Hate speech detection includes a lot of aspects: misogyny, sexism, racism, aggression, violence, toxicity, etc. The overview of the existing hate speech results and solutions is presented in~\cite{info13060273}.
The task of hate speech detection can be made more feasible if specific hate speech types are considered, and for that, multiple datasets and solutions have been designed. Among such solutions, there are neural network-based models~\cite{DBLP:conf/sepln/GoenagaAGCIEOPP18a}, a combination of different simple classification algorithms~\cite{DBLP:conf/sepln/ShushkevichC18}, and document embeddings with classical models~\cite{DBLP:conf/sepln/0002CC18}. More specifically, in this section we discuss different approaches and datasets to hate speech detection for Spanish as our case study. In IberEval @SEPLN 2018, \cite{fersini2018overview} proposed the Automatic Misogyny Identification (AMI) task. It consists of misogyny identification, misogynistic behavior categorization, and target classification, using Spanish and English Twitter messages. 
\citep{anzovino2018automatic} presented a corpus of misogynous tweets and an exploratory investigation of machine learning (ML) models for detecting and classifying misogynistic language.
In~\cite{10.1145/3369869}, a report on hate speech research in Spanish tweets against women and immigrants is presented. The authors compared different algorithms and created language resources for hate speech detection in Spanish.
In~\cite{tonja2022detection}, a work for the shared task @IberLEF2022 was presented that examines the use of a language-specific pre-trained language model for tackling the issue of detection of aggressive and violent social media texts in Spanish.

The problem of sexism detection as a specific case of hate speech was investigated in~\cite{ranasinghe-zampieri-2020-multilingual}. The authors worked with an available English dataset by using cross-lingual contextual word embeddings and transfer learning techniques to predict sexism in a multi-lingual setup (including Spanish). 
\cite{https://doi.org/10.48550/arxiv.2111.04551} used multilingual and monolingual BERT algorithms, data point translation, and ensemble strategies for sexism detection in English and Spanish. 
\cite{VALLECANO2023119446} presented the SocialHaterBERT model to detect hate speech using social media data. They fine-tuned the BETO-based model, \cite{CaneteCFP2020}, on hate speech tweets in Spanish. The authors used a transformer model that outperformed previous solutions.
In~\cite{hasoc2020} challenge the offensive hate speech dataset was presented that included insulting, derogatory or obscene, and hurtful content in social media.

\section*{Task}\label{task}\hypertarget{sec:task}{}

\begin{figure}[!htb]
    \centering
    \includegraphics[width=\linewidth]{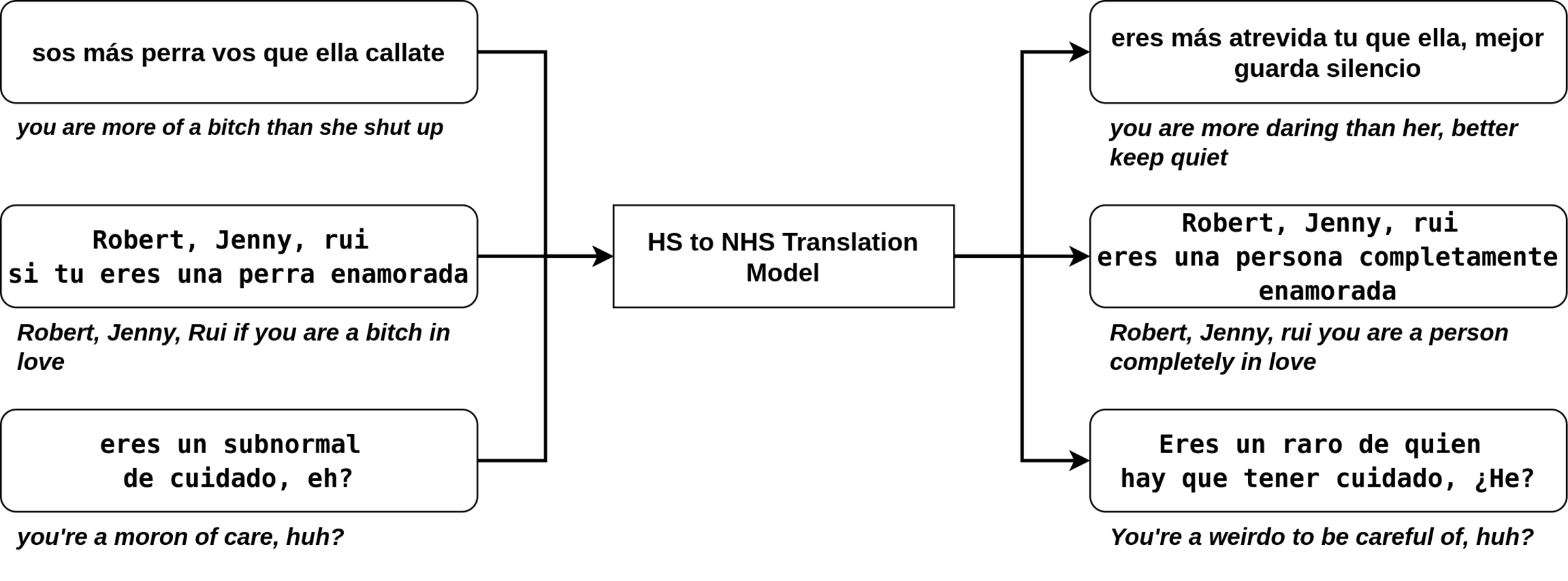}
    \caption{Hate speech to non-hate speech translation task}
    \label{fig:hs2nhs}
\end{figure}

We propose a new task of \textit{hate speech translation}. The main idea of the task is to build an algorithm that rewrites hate speech text into non-hate speech by keeping the text semantics intact. To illustrate, we provide an example of a hate speech text (HS) from our dataset and its possible non-hate speech translations (NHS), with English translations (see Figure~\ref{fig:hs2nhs}).

The algorithm we propose takes as input a hate speech sample and generates an appropriate non-hate speech text, which is a softened and non-hate version of the input. More formally, the algorithm can be described as follows. 
Let $x$ be the input hate speech text represented as a sequence of tokens, and let $y$ be the output non-hate speech text, also represented as a sequence of tokens. We want to generate $y$ given $x$ by modeling the conditional distribution $p(y|x)$. A probabilistic language generation model estimates this distribution by learning a set of parameters $\theta$ that maximize the likelihood of observing the training data. The objective is to maximize the log-likelihood of the training set (see Equation~\ref{eq:likel}):
\begin{equation}
    \sum_{i=1}^{N}\log p(y^{(i)}|x^{(i)}; \theta) \label{eq:likel}
\end{equation}

where $N$ is the size of the training set, and $y^{(i)}$ and $x^{(i)}$ are the input and output sequences for the $i$-th training example.

One common approach to modeling $p(y|x)$ is the sequence-to-sequence architecture, which consists of an encoder and a decoder neural network. The encoder takes the input sequence $x$ and produces a fixed-length vector representation, which is then used as input to the decoder network to generate the output sequence $y$. During training, the decoder network generates the output sequence $y$ token by token, conditioning on the $x$ and the previously generated tokens of $y$. The model estimates the probability of each token in the output sequence given the input sequence and the previously generated tokens and selects the most likely token at each step. At inference time, the model generates the output sequence by sampling from the conditional distribution $p(y|x)$ using beam search or another decoding algorithm.

Hate speech translation is an important task because it has the potential to create a more inclusive and respectful society, as well as to make social media a safer place. Hate speech has a harmful impact on individuals and groups, spreading prejudice and division. By translating hate speech into non-hate speech, we can help to counter these negative effects, promote a more positive public discourse, and improve the overall way of online communication. This task has also the potential to support research and advance in the field of natural language processing, as well as in the study of hate speech and its impacts. To sum it up, work on hate speech translation is valuable in terms of both its practical applications and its contributions to our understanding of language and society.

\section*{Dataset}\label{dataset}\hypertarget{sec:dataset}{}

To create a dataset for our task, we used the Spanish hate speech dataset from IberEval 2018~\cite{DBLP:conf/sepln/2018ibereval} as a base. Hashtags, mentions, usernames, URL links, and email addresses were removed from the texts. To create a hate speech translation dataset, a group of seven native Spanish-speaking Mexican students was recruited. For each hate speech text, the students were asked to provide at least three examples of the same text without the hate speech, while retaining the original sense of the text. If the hate speech contained only insults without any discernible meaning, the students were instructed to skip it. These samples were not included in the dataset. To ensure the diversity and reliability of the hate speech translations, the examples were randomly assigned to different students, and it was rare for different students to label the same example in the same way. To label the data we used doccano~\cite{doccano}. 

The decision to ask for three examples per text was made based on the need for diversity in translations. Having multiple translations allows for a broader range of non-hateful expressions to be included, which can help improve the performance and validation of the hate speech translation models. Additionally, multiple translations can provide insights into the different ways in which people might express similar ideas without resorting to hate speech. 

It should be noted that due to time and resource constraints, not all texts in the base dataset from IberEval 2018~\cite{DBLP:conf/sepln/2018ibereval} were translated. Moreover, we discovered that 41 English texts appeared in the dataset, but people still labeled them by providing translations in Spanish, with a total of 120 translations. While this represents only 2\% of all data and is therefore relatively insignificant, it is important to acknowledge their presence and potential impact on the dataset.

The total number of hate speech texts in the final datasets is 1,680. The total number of translations equals 5,208. Some examples from our dataset are presented in Table~\ref{tab:example_data}. English translations are presented in Table~\ref{tab:example_data_en}


    

\begin{table}[h!]
\begin{tabular}{|c|c|}
\hline
Spanish Hate Speech                                                                                                                      & Spanish Non-Hate Translations                                                                 \\ \hline
\makecell{en que se parece una mujer a un cientifico?\\en que al cientifico le pasan\\cosas por la cabeza,\\y a la mujer le pasan cabezas por la cosa} & \makecell{1. ¿En que se pare una mujer a un cient\'{i}fico?\\En que a ambos les pasan cosas por la cabeza.}    \\ \cline{2-2} 
& \makecell{2. La similitud entre un cient\'{i}fico y una mujer\\es que a ambos les entran cosas por la cabeza.} \\ \cline{2-2} 
& \makecell{3. La mujer se parece a un cient\'{i}fico\\al entrarles pensamientos por la cabeza.}                \\ \hline
\makecell{Gloria te mereces todo lo que\\est\'{a} bien, que lindo verte as\'{i} perra}                       & \makecell{1. Gloria te mereces todo lo que est\'{a} bien,\\que lindo verte as\'{i} guapa} \\ \cline{2-2} 
 & \makecell{2.  Gloria te mereces todo lo que est\'{a} bien,\\que lindo verte as\'{i} hermosa}       
 \\ \cline{2-2} 
& \makecell{3. Gloria te mereces todo lo que est\'{a} bien,\\que lindo verte as\'{i} preciosa}                       \\ \hline
\makecell{ya callate puta perra me tienes cansado \\con tus pendejadas}                               & \makecell{1. Podr\'{i}as guardar silencio me tienes\\cansado con tus cosas }  \\ \cline{2-2} 
 & \makecell{2. Ya guarda silencio,\\me tienes cansado con tus bober\'{i}as}                                      \\ \cline{2-2} 
& \makecell{3. Ya deja de hablar,\\estoy cansado de tus cosas}                                             \\ \hline
\makecell{Hattie, Debra, lo unico que vas a tener en\\tu culo es mi polla}                          & \makecell{1. Hattie, Debra, no tendr\'{a}s nada. }                                                            \\ \cline{2-2} 
& \makecell{2. Hattie, Debra, Te quedaras sin nada.}                                                        \\ \cline{2-2} 
 & \makecell{3. Hattie, Debra, no te dar\'{e} nada.}  \\ \hline
\end{tabular}
    \caption{Examples from our dataset}
    
    \label{tab:example_data}
\end{table}

\begin{table}[h!]
\begin{tabular}{|c|c|}
\hline
English Translation of Hate Speech                                                                                                         & English Translation of Non-Hate Translations                                                         \\ \hline
\makecell{How is a woman like a scientist?\\in which the scientist has things\\ going through his head,\\and the woman has heads\\going through the thing} & \makecell{1. What does a woman stop a scientist?\\In which both have things going through their heads. }          \\ \cline{2-2} 
   & \makecell{2. The similarity between a scientist\\ and a woman is that both\\ have things going through their heads.} \\ \cline{2-2} 
  & \makecell{3. The woman looks\\like a scientist \\when thoughts enter her head.}                                     \\ \hline
\makecell{Gloria you deserve everything\\that is right,\\how nice to see you\\like this bitch }                                                 & \makecell{1. Gloria you deserve everything\\ that is good, \\how nice to see you so beautiful}                       \\ \cline{2-2} 
   & \makecell{2. Gloria you deserve everything\\that is good,\\how nice to see you so beautiful  }                     \\ \cline{2-2} 
   & \makecell{3. Gloria you deserve everything\\that is good,\\how nice to see you\\like this precious }                \\ \hline
\makecell{shut up bitch bitch\\you have me tired\\with your bullshit  }                                                                                 & \makecell{1. Could you keep quiet\\ you have me tired with your things }                                           \\ \cline{2-2} 
   & \makecell{2. Be silent, you have me \\tired with your nonsense    }                                                \\ \cline{2-2} 
   & \makecell{3. Stop talking,\\I'm tired of your things    }                                                         \\ \hline
\makecell{Hattie, Debra, the only thing \\you're going to get in \\your ass is my dick.}                                                                  & \makecell{1. Hattie, Debra, \\you won't have anything.      }                                                      \\ \cline{2-2} 
   & \makecell{2. Hattie, Debra, you'll \\be left with nothing.   }                                                     \\ \cline{2-2} 
   & \makecell{3. Hattie, Debra,\\I won't give you anything.  }                                                        \\ \hline
\end{tabular}

    \caption{Examples from our dataset translated to English}
    
    \label{tab:example_data_en}
\end{table}

\begin{figure}[!htb]
    \centering
    \includegraphics[width=\linewidth]{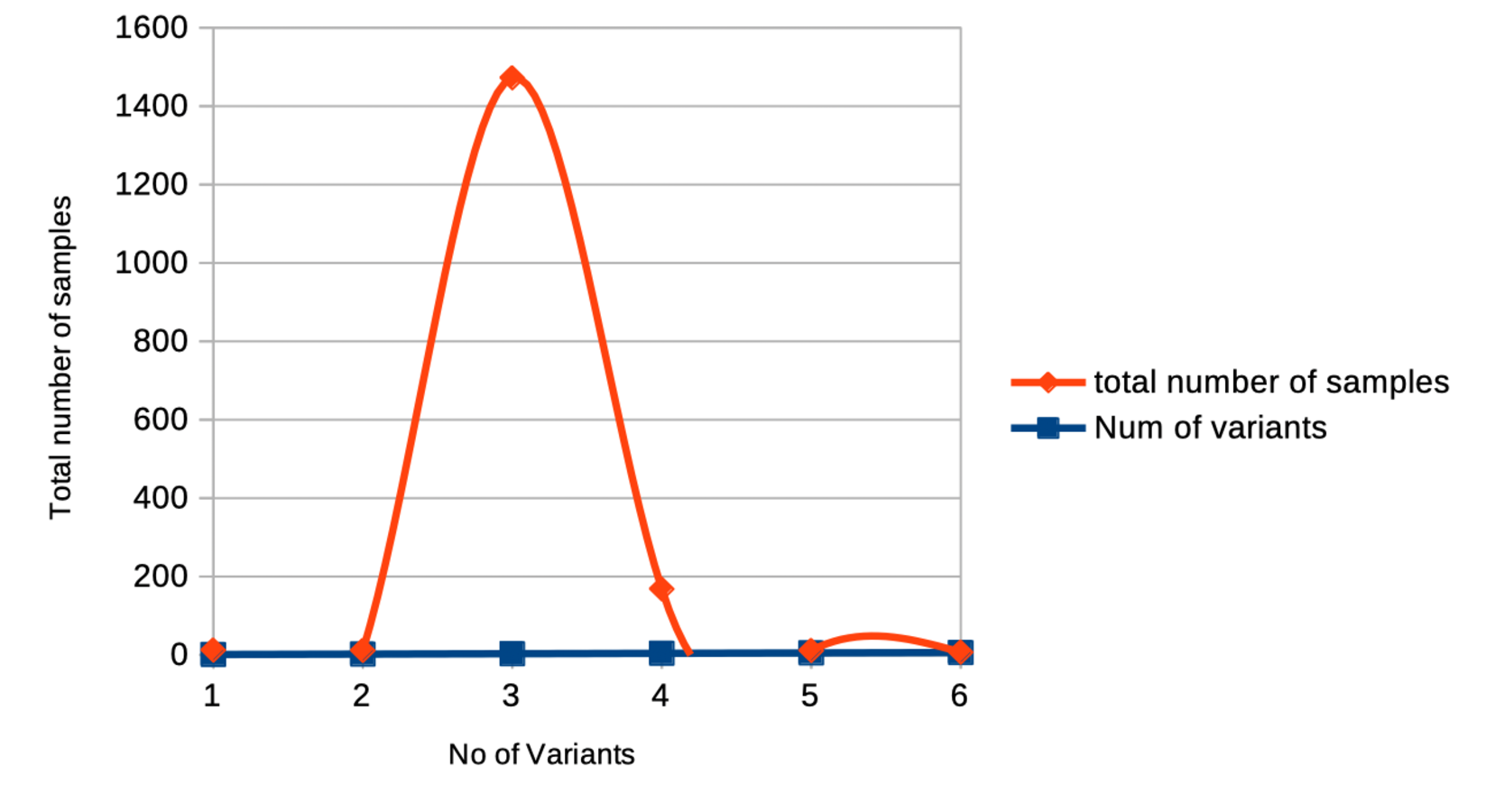}
    \caption{Distribution of the number of translations per text}
    \label{fig:NumTransDistr}
\end{figure}

The distribution of the number of non-hate variants is shown in Figure~\ref{fig:NumTransDistr}. The average number of non-hate speech translations per text is 3.1 and the median is 3.0.



\section*{Baseline Methodology}\hypertarget{sec:method}{}
For our translation task, we choose MT models trained to translate one language to another because our dataset contains one-to-many translation parallel data in the same (Spanish) language. We used four different widely used MT models for our experiments, namely the RNN-based encoder-decoder model, T5-small model, mBART-based model, and Marian-based model:

\begin{itemize}
    \item RNN-based model is based on the sequence-to-sequence architecture and uses an RNN-based encoder to encode the input sequence and a decoder with an attention mechanism to generate the output sequence. While this model is not as advanced as the transformer-based models, it still performs well on various NLP tasks and can generate fluent and coherent translations.
    \item T5 small model~\cite{2020t5} is a transformer-based text-to-text model. Every task (or prompt) is framed as feeding a text as input into the model and training it to output some target value, including translation, the answer to the question, or the category of the text.
    \item mBART-based~\cite{mbart} model is a sequence-to-sequence architecture. Specifically, we used \textit{mBART-large}~\footnote{\url{https://huggingface.co/facebook/mbart-large-50-many-to-many-mmt}}, which was fine-tuned for the multilingual machine translation task. mBART was originally pre-trained using the BART~\cite{lewis-etal-2020-bart} objective on datasets in numerous languages. A single Transformer model was learned to recover the texts from noised input texts that were created by masking phrases and permuting sentences.
    \item Marian-based~\cite{mariannmt} model is another sequence-to-sequence architecture we used in our experiments. Specifically, we used an OPUS-trained~\cite{opus} translation model for English to Spanish~\footnote{\url{https://huggingface.co/Helsinki-NLP/opus-mt-en-es}}, which was introduced by Helsinki-NLP group~\cite{TiedemannThottingal:EAMT2020}.
\end{itemize}

We selected the above models based on their ability to handle different types of text inputs and outputs, their performance on various NLP tasks, and their suitability for our specific hate speech translation task in Spanish. The T5 small model is suitable for handling short text inputs and outputs, while the mBART-based model handles multilingual inputs and outputs. The Marian-based model is suitable for handling complex sentence structures and high-quality translations. We fine-tuned each of these models on our hate speech dataset.

\section*{Testing and Evaluation Methodology}\label{eval}\hypertarget{sec:eval}{}
\subsection*{Data Split}

As our dataset is considered to be a low-resource dataset, we used the k-fold validation technique to split the data into train and test chunks. We split our data into four folds based on the hate speech samples. It means that if a hate speech sample is assigned to the train set, then all of its translations are assigned there as well. The splits statistics are presented in Table~\ref{tab:folds}. The obtained folds are uniform in the number of train and test samples.

\begin{table}[!htb]
\centering
\begin{tabular}{|p{0.1\textwidth}|p{0.1\textwidth}p{0.1\textwidth}|p{0.1\textwidth}p{0.1\textwidth}|}
\hline
\makecell{Fold\\number} & \multicolumn{2}{l|}{Train}                                                 & \multicolumn{2}{l|}{Test}                                                 \\ 
            & \multicolumn{1}{p{0.1\textwidth}|}{Num. of hate speech texts} & Total num. of samples & \multicolumn{1}{p{0.1\textwidth}|}{Num. of hate speech texts} & Total num. of samples \\ \hline
1           & \multicolumn{1}{l|}{1,260}                        & 3,910                    & \multicolumn{1}{l|}{420}                         & 1,298                    \\ \hline
2           & \multicolumn{1}{l|}{1,260}                        & 3,899                    & \multicolumn{1}{l|}{420}                         & 1,309                    \\ \hline
3           & \multicolumn{1}{l|}{1,260}                        & 3,896                    & \multicolumn{1}{l|}{420}                         & 1,312                    \\ \hline
4           & \multicolumn{1}{l|}{1,260}                        & 3,919                    & \multicolumn{1}{l|}{420}                         & 1,289                    \\ \hline
\end{tabular}
\caption{Folds` statistics}
\label{tab:folds}
\end{table}

\subsection*{Metrics}

To evaluate the quality of the translation algorithm, we used several BLEU-based metrics from the NLTK~\cite{bird2009natural} package, which is one of the known machine translation metrics to evaluate the performance of translation models.

As aggregate measures, the mean, median, and variance were selected. The mean and the median give an idea of the average performance of the model across all the folds, while the variance indicates the spread of the results. A lower variance implies that the performance of the model is more consistent across different folds, while a higher variance indicates more variability in the results.
The BLEU-based metrics we used are as follows.

\begin{itemize}
    \item Sentence BLEU Score: it is the BLEU score computed for each sentence in the evaluation set. 
    \item N-grams Sentence BLEU Score: it is Sentence BLEU Score, but to compute the score, it takes into account N-grams instead of single words. The values of N we used were 2 and 3.
\end{itemize}

\section*{Experiments and Results}\label{exper}\hypertarget{sec:exper}{}

The average results per fold are presented in Table~\ref{tab:avg_results}.

\begin{table}[!htb]
\centering
\begin{tabular}{|c|c|c|c|c|}
\hline

\textbf{Metric} & \textbf{RNN} & \textbf{T5} & \textbf{mBART} & \textbf{Marian} \\\hline

mean\_sentence\_bleu\_scores & 0.022 & 0.177 & \textbf{0.236} & 0.184 \\\hline
mean\_three\_grams\_sentence\_bleu\_scores & 0.024 & 0.186  &  \textbf{0.244}  &  0.193    \\\hline
mean\_two\_grams\_sentence\_bleu\_scores &  0.046  & 0.247 &  \textbf{0.313}    &  0.267      \\\hline
median\_sentence\_bleu\_scores         &   0.000  & 0.012  &   \textbf{0.141}    &   0.018    \\\hline
median\_three\_grams\_sentence\_bleu\_scores& 0.000 & 0.105 &  \textbf{0.170}      &  0.114 \\\hline
median\_two\_grams\_sentence\_bleu\_scores & 0.000 & 0.184  &  \textbf{0.263}  &  0.223    \\\hline
var\_sentence\_bleu\_scores              &  \textbf{0.008} & 0.059   &  0.075  &  0.059   \\\hline
var\_three\_grams\_sentence\_bleu\_scores & \textbf{0.007}  & 0.053  & 0.067     & 0.052   \\\hline
var\_two\_grams\_sentence\_bleu\_scores   &  \textbf{0.013}  & 0.058 & 0.069     & 0.057 \\\hline

\end{tabular}
\caption{Average results on folds}
\label{tab:avg_results}
\end{table}

The mean sentence BLEU scores show that the best-performing model is mBART, with a score of 0.236. The mean 2-gram and 3-gram sentence BLEU scores also show the best performance by mBART, with a score of 0.313 and 0.244 respectively. 

The median sentence BLEU scores indicate that T5 performed the worst, with a score of 0.012. The median 2-gram sentence BLEU scores show the best performance by mBART, with a score of 0.263. The median 3-gram sentence BLEU scores show a similar pattern, with mBART performing the best with a value of 0.17.

The variance of sentence BLEU scores and the variance of n-gram sentence BLEU scores indicate the variability in the results, with smaller values showing more consistent results. T5, mBART, and Marian models have similar variances in both categories, with mBART showing the biggest variance in sentence BLEU scores. THe RNN model has the smallest variances.

The RNN model showed the worst performance compared to the transformers-based models among all of the metrics due to the simplicity of the architecture.

In conclusion, the results show that the mBART model performed the best overall, with the highest mean and median sentence BLEU scores, as well as the highest median n-gram sentence BLEU scores. 

The obtained results for each fold are presented in Tables~\ref{tab:folds_rnn},~\ref{tab:folds_t5},~\ref{tab:folds_mbart}, and~\ref{tab:folds_marian}.

\begin{table}[!htb]
\centering
\begin{tabular}{|c|c|c|c|c|}
\hline

\textbf{Metric}                              & \textbf{Fold 1} & \textbf{Fold 2} & \textbf{Fold 3} & \textbf{Fold 4} \\\hline
mean\_sentence\_bleu\_scores                 & 0.025          & 0.018          & 0.020          & 0.024          \\\hline
mean\_three\_grams\_sentence\_bleu\_scores   & 0.027          & 0.020          & 0.022          & 0.026          \\\hline
mean\_two\_grams\_sentence\_bleu\_scores     & 0.053          & 0.042          & 0.044          & 0.047          \\\hline
median\_sentence\_bleu\_scores               & 0.000          & 0.000          & 0.000          & 0.000          \\\hline
median\_three\_grams\_sentence\_bleu\_scores & 0.000          & 0.000          & 0.000          & 0.000          \\\hline
median\_two\_grams\_sentence\_bleu\_scores   & 0.000          & 0.000          & 0.000          & 0.000          \\\hline
var\_sentence\_bleu\_scores                  & 0.010          & 0.006          & 0.007          & 0.010          \\\hline
var\_three\_grams\_sentence\_bleu\_scores    & 0.009          & 0.006          & 0.006          & 0.009          \\\hline
var\_two\_grams\_sentence\_bleu\_scores      & 0.014          & 0.010          & 0.012          & 0.014          \\ \hline

\end{tabular}
\caption{RNN model results}
\label{tab:folds_rnn}
\end{table}

\begin{table}[!htb]
\centering
\begin{tabular}{|c|c|c|c|c|}
\hline

\textbf{Metric}                              & \textbf{Fold 1} & \textbf{Fold 2} & \textbf{Fold 3} & \textbf{Fold 4} \\\hline
mean\_sentence\_bleu\_scores                 & 0.177           & 0.174           & 0.192           & 0.166                      \\\hline
mean\_three\_grams\_sentence\_bleu\_scores   & 0.185           & 0.187           & 0.195           & 0.177                      \\\hline
mean\_two\_grams\_sentence\_bleu\_scores     & 0.249           & 0.250           & 0.255           & 0.234                      \\\hline
median\_sentence\_bleu\_scores               & 0.000           & 0.000           & 0.048           & 0.000                      \\\hline
median\_three\_grams\_sentence\_bleu\_scores & 0.097           & 0.108           & 0.114           & 0.099                      \\\hline
median\_two\_grams\_sentence\_bleu\_scores   & 0.182           & 0.184           & 0.200           & 0.170                      \\\hline
var\_sentence\_bleu\_scores                  & 0.063           & 0.059           & 0.061           & 0.052                      \\\hline
var\_three\_grams\_sentence\_bleu\_scores    & 0.057           & 0.052           & 0.055           & 0.047                      \\\hline
var\_two\_grams\_sentence\_bleu\_scores      & 0.061           & 0.058           & 0.059           & 0.053                    \\ \hline

\end{tabular}
\caption{T5 model results}
\label{tab:folds_t5}
\end{table}

\begin{table}[!htb]
\centering
\begin{tabular}{|c|c|c|c|c|}
\hline
\textbf{Metric}                              & \textbf{Fold 1} & \textbf{Fold 2} & \textbf{Fold 3} & \textbf{Fold 4} \\\hline
mean\_sentence\_bleu\_scores                 & 0.230          & 0.249          & 0.237          & 0.228          \\\hline
mean\_three\_grams\_sentence\_bleu\_scores   & 0.236          & 0.256          & 0.250          & 0.233          \\\hline
mean\_two\_grams\_sentence\_bleu\_scores     & 0.311          & 0.321          & 0.320          & 0.299          \\\hline
median\_sentence\_bleu\_scores               & 0.111          & 0.154          & 0.167          & 0.131          \\\hline
median\_three\_grams\_sentence\_bleu\_scores & 0.143          & 0.195          & 0.192          & 0.150          \\\hline
median\_two\_grams\_sentence\_bleu\_scores   & 0.250          & 0.279          & 0.273          & 0.250          \\\hline
var\_sentence\_bleu\_scores                  & 0.079          & 0.077          & 0.073          & 0.072          \\\hline
var\_three\_grams\_sentence\_bleu\_scores    & 0.071          & 0.067          & 0.066          & 0.064          \\\hline
var\_two\_grams\_sentence\_bleu\_scores      & 0.072          & 0.070          & 0.067          & 0.068             \\\hline

\end{tabular}
\caption{mBART model results}
\label{tab:folds_mbart}
\end{table}

\begin{table}[!htb]
\centering
\begin{tabular}{|c|c|c|c|c|}
\hline
\textbf{Metric}                              & \textbf{Fold 1} & \textbf{Fold 2} & \textbf{Fold 3} & \textbf{Fold 4} \\\hline
mean\_sentence\_bleu\_scores                 & 0.175          & 0.200          & 0.184          & 0.177          \\\hline
mean\_three\_grams\_sentence\_bleu\_scores   & 0.183          & 0.208          & 0.192          & 0.189          \\\hline
mean\_two\_grams\_sentence\_bleu\_scores     & 0.262          & 0.281          & 0.261          & 0.262          \\\hline
median\_sentence\_bleu\_scores               & 0.000          & 0.071          & 0.000          & 0.000          \\\hline
median\_three\_grams\_sentence\_bleu\_scores & 0.096          & 0.125          & 0.105          & 0.130          \\\hline
median\_two\_grams\_sentence\_bleu\_scores   & 0.202          & 0.250          & 0.205          & 0.236          \\\hline
var\_sentence\_bleu\_scores                  & 0.059          & 0.062          & 0.063          & 0.053          \\\hline
var\_three\_grams\_sentence\_bleu\_scores    & 0.052          & 0.054          & 0.055          & 0.048          \\\hline
var\_two\_grams\_sentence\_bleu\_scores      & 0.058          & 0.058          & 0.059          & 0.055         \\\hline

\end{tabular}
\caption{Marian model results}
\label{tab:folds_marian}
\end{table}

From the results, it can be seen that the T5 model performs relatively well in translating hate speech into non-hate speech in Spanish. The mean 2-gram sentence BLEU scores are the highest among the metrics, with values ranging from 0.234 to 0.255 across the four folds. The median 2-gram sentence BLEU scores are also relatively high, with values ranging from 0.170 to 0.200. This indicates that the model generates non-hate speech translations that have a good overlap with the reference translations in terms of 2-gram.

The results of mBART model indicate that, overall, the model's performance is better than that of the other models. Mean sentence BLEU scores range from 0.228 to 0.249 and the mean 2-gram sentence BLEU scores range from 0.299 to 0.321. The variance in sentence BLEU scores is higher than for the others models, ranging from 0.071 to 0.079. 

The Marian model performs consistently across all four folds. The mean and median sentence BLEU scores have the values with the highest mean score being 0.200 and the highest median score being 0.281. The variance of sentence BLEU scores is relatively low, with a range of 0.053 to 0.063.

The output for the models trained on the first fold is presented in Tables~\ref{tab:span_preds} and~\ref{tab:en_preds}. In Table~\ref{tab:span_preds} the original Spanish translations are shown and the English-translated version is presented in Table~\ref{tab:en_preds}.

\begin{table}[]
\begin{tabular}{|c|c|l|}
\hline
\textbf{Input Text}                   & \textbf{Model} & \textbf{Prediction}                                                                                  \\ \hline
\makecell{Delores, Kimberly,\\ c\'allate tu !! perra}     & RNN             & \makecell{Rusell , Hilary ,\\ya no digas nada por favor !}                      \\ \cline{2-3}
                                        & T5             & Delores, Kimberly, c\'{a}llate tu!!                                                                       \\ \cline{2-3} 
                                      & mBART          & Delores, Kimberly, guarda silencio!!                                                                        \\ \cline{2-3} 
                                      & Marian         & \makecell{Espero que te quedes$\_$callada,\\ te quedas sin verg\"uenza.}\\ \hline
Michael pa' pegarte en la cara marica & RNN             & \makecell{Marquita que que es mejor ,\\est\'{a} no seas bien .\\y ojal\'{a} y no es .}          \\ \cline{2-3} 
                                        & T5             & Michael pa en la cara marica.                                                                      \\ \cline{2-3} 
                                      & mBART          & Michael ya no hables mas                                                                         \\ \cline{2-3} 
                                      & Marian         & Michael pa' $\_$pegarte en la cara.                                                                \\ \hline
David tu eres mi hermana perra         & RNN             & tu eres mi hermana , eres mi hermana         \\ \cline{2-3} 
                                    & T5             & David tu eres mi hermana perra                                                                        \\ \cline{2-3} 
                                      & mBART          & David eres mi hermana                                                                              \\ \cline{2-3} 
                                      & Marian         & David eres mi hermana                                                                              \\ \hline
\makecell{Deborah pero voy como\\ a las 12:30 por ahi, \\puta zorra no te mereces que\\ te lleve pizzas}     & RNN             & \makecell{no mereces y lo que no , te no . }                                                                       \\ \cline{2-3} 
                                 & T5             & \makecell{Deborah pero voy\\ como a las 12:30 por ahi}                                \\ \cline{2-3} 
                                      & mBART          & \makecell{Deborah pero voy\\ como a las 12:30 por ahi, no}                                                                             \\ \cline{2-3} 
                                      & Marian         & \makecell{$\_$Deborah pero voy como\\ a las 12:30 por ahi, no}                                                                              \\ \hline
\end{tabular}
\caption{Predictions of different models in Spanish}
\label{tab:span_preds}
\end{table}

\begin{table}[]
\begin{tabular}{|c|c|l|}
\hline
\textbf{Input Text}                   & \textbf{Model} & \textbf{Prediction}                                                                                  \\ \hline

Delores, Kimberly, shut up! bitch     & RNN             & \makecell{Russell, Hilary,\\please don't say anything anymore!}                                                   \\ \cline{2-3}
                                        & T5             & Delores, Kimberly, shut up!!                                                                         \\ \cline{2-3} 
                                      & mBART          & Delores, Kimberly, be quiet!!                                                                        \\ \cline{2-3} 
                                      & Marian         & \makecell{I hope you stay quiet, \\you're shameless.} \\ \hline
Michael to hit you in the face fagot. & RNN             & \makecell{Marquita that is better,\\you are not well.\\and hopefully and it is not.}                           \\ \cline{2-3} 
                                        & T5             & Michael pa in the sissy face.                                                                        \\ \cline{2-3} 
                                      & mBART          & Michael don't talk anymore                                                                           \\ \cline{2-3} 
                                      & Marian         & Michael to $\_$hit you in the face.                                                                  \\ \hline
David you are my sister bitch         & RNN             & \makecell{you are my sister,\\you are my sister}                                                                        \\ \cline{2-3} 
                                    & T5             & David you are my sister bitch                                                                        \\ \cline{2-3} 
                                      & mBART          & David you are my sister                                                                              \\ \cline{2-3} 
                                      & Marian         & David you are my sister                                                                              \\ \hline
\makecell{Deborah but I'm going like\\ at 12:30 out there, \\bitch bitch you don't deserve me to\\ bring you pizzas}        & RNN             & \makecell{you don't deserve and\\what you don't,\\you don't.} \\\cline{2-2}

                                        & T5             & \makecell{Deborah, but I'm going\\ around 12:30.}    \\\cline{2-3} 
                                      & mBART          & \makecell{Deborah, but I'm going\\ around 12:30, no}                                                                              \\ \cline{2-3} 
                                      & Marian         & \makecell{$\_$Deborah, but I'm going like\\ at 12:30 there, no}                                                                              \\ \hline
\end{tabular}
\caption{Predictions of different models translated to English}
\label{tab:en_preds}
\end{table}

In the first example, the input text contains an insult directed towards someone named Delores and Kimberly. The T5 model correctly detected the insult, and it removed the insulting word "perra" (``bitch'') at the end. The mBART model changed the insulting word ``perra'' to ``guarda silencio'' (``keep quiet''). The Marian model predicted a response in a different form that contains an indirect insult, ``Espero que te quedes callada, te quedas sin verg\"uenza'' (``I hope you stay quiet, you're shameless''). In this example, mBART showed better results, as it could soften the language significantly better than the other models. The Marian model used an indirect insult, which still can be considered hate speech in some cases. The T5 model removed the obvious hateful word, but at the same time, the sentiment of the output remained insulting. The RNN model successfully translated the hate speech phrase to ``ya no digas nada por favor !'' (``please don't say anything now.''). However, the model generated different names as replacements, which suggests that the model may not have fully captured the context of the input and made some errors in the name replacement. The problem also may occur due to the vocabulary limitations of the RNN model: if the word is not present in the vocabulary, then the model will not be able to generate it.

In the second example, a threat towards someone named Michael was presented as input. The T5 model detects the threat, but it does not use the word ``pegarte'' (``hit you''). Instead, it predicted ``Michael pa en la cara marica'' (``Michael in the face, sissy.''). The mBART model generated a response telling Michael to stop talking. The Marian model generated a response that included the word ``pegarte'' (``hit you''). In this example, the mBART removed the negative sentiment from the sentence, but the other models did not. The RNN model translated the hate speech phrase to ``Marquita que que es mejor , está no seas bien . y ojalá y no es .'' (``Marquita that is better,you are not well.and hopefully and it is not.''). It seems to be a mix of different phrases. It suggests that the model may have struggled with the specific vocabulary used in the input and may need more training data or a more sophisticated architecture to improve its performance.

In the third example, the input text is an insult directed towards someone named David, calling him a ``perra'' (``bitch''), and stating that he is the speaker's sister. The T5 and Marian models generated the same response, very similar to the input text, while the mBART model omitted the insulting word ``perra'' (``bitch''). All of the models struggled with this example, as there is no additional sense provided to it, only pure insult. From the text, we can conclude that David is not a sister of the speaker. The input text did not have any more, its only purpose was to call David names. The RNN model generated ``tu eres mi hermana , eres mi hermana'' (``you are my sister, you are my sister''). It does not provide a meaningful translation. The model may have encountered difficulties in replacing certain words with more appropriate alternatives, and more data and training are needed to improve its performance.

In the fourth example, the input text states that the speaker will arrive around 12:30 and insult someone called Deborah by calling her a ``puta zorra'' (``whore''). The T5 model generated a response that included the time of arrival and omitted the insult. The mBART model generated a response that included the time of arrival and removed the insult by saying "no" at the end. The Marian model generated a response that also includes the time of arrival and omitted the insult. The RNN model's result is ``no mereces y lo que no , te no .'' (``you don't deserve and what you don't,\\you don't.''), which is not grammatically correct and does not maintain the same meaning as the original input. This suggests that the model has encountered difficulties in handling the complex sentence structure and the specific vocabulary used in the input.

Overall, the three models generated different responses to the same input texts, demonstrating different approaches to text processing. Concerning the metrics in
Tables~\ref{tab:avg_results},~\ref{tab:folds_rnn},~\ref{tab:folds_t5},~\ref{tab:folds_mbart}, and ~\ref{tab:folds_marian}, the mBART showed the best results.



\section*{Conclusions}\label{conc}\hypertarget{sec:conc}{}

Translation of hate speech into non-hate speech is a crucial task in today's society. With the increasing spread of hate speech through various media platforms, it has become essential to fight the impact of such speech and promote positive and inclusive communication. The results of our study highlight the potential of deep learning techniques for translating hate speech into non-hate speech. However, the limitations of the dataset we gathered and applied and the use of simple baselines demonstrate the need for further research and improvement in this area. Future studies can explore the use of more advanced deep learning models, and bigger and more diverse datasets. The development of more accurate and robust models for this task will have a positive impact on the reduction of hate speech and the promotion of positive and inclusive communication. Thus, we believe that this task opens an important area of research and deserves the attention of the NLP community.

\section*{Acknowledgments}

The work was done with support from the Mexican Government through the grant A1-S-47854 of CONACYT, Mexico, and two grants of the Secretaría de Investigación y Posgrado of Instituto Politécnico Nacional, Mexico. The authors thank the CONACYT for the computing resources brought to them through the Plataforma de Aprendizaje Profundo para Tecnologías del Lenguaje of the Laboratorio de Supercómputo of the INAOE, Mexico and acknowledge the support of Microsoft through the Microsoft Latin America PhD Award.

\bibliography{sample}

\end{document}